\documentclass[letterpaper,journal]{IEEEtran}
\usepackage{amsmath,amsfonts}
\usepackage{algorithmic}
\usepackage{algorithm}
\usepackage{array}
\usepackage[caption=false,font=normalsize,labelfont=sf,textfont=sf]{subfig}
\usepackage{textcomp}
\usepackage{stfloats}
\usepackage{url}
\usepackage{verbatim}
\usepackage{graphicx}
\usepackage{booktabs}
\usepackage{makecell}
\usepackage[figuresleft]{rotating}

\usepackage{multirow}
\usepackage[nocompress]{cite}
\begin{document}

\title{GlassGuard: Verified Glass Plane Mapping for Robot Navigation}

\author{Hanwen Guo$^{1}$, Zhengzhi Lin$^{1}$, Yusen Xie$^{1}$, and Ji Zhang$^{1}$
\thanks{$^{1}$The authors are with Carnegie Mellon University, Pittsburgh, PA,
USA.}
\thanks{This work has been submitted to the IEEE for possible publication.
Copyright may be transferred without notice, after which this version may no
longer be accessible.}}

\markboth{IEEE Robotics and Automation Letters. Preprint version.}%
{Guo \MakeLowercase{\textit{et al.}}: GlassGuard}
\pagestyle{empty}

\maketitle
\thispagestyle{empty}

\begin{abstract}
Transparent and specular surfaces pose a serious challenge to LiDAR-based SLAM and navigation because laser returns may pass through glass, leaving collision boundaries absent from the map. Prior work attempts to reconstruct the missing surfaces, but inaccurate obstacle placement can create the opposite failure: contamination of traversable free space. Recognizing this dual requirement, we present GlassGuard, a navigation-oriented framework for reconstructing planar architectural glass from complementary visual and LiDAR evidence. We formulate success in terms of both glass coverage and free-space contamination and apply this principle throughout proposal verification and global map construction. A foundation vision model provides glass-instance masks, structural 3D cues generate metric plane hypotheses, and depth-free 2D projective geometry checks their orientations before they enter a consolidated global map. We evaluate GlassGuard in nine building-scale scenes spanning diverse glass structures, spatial scales, and lighting conditions, with more than one hour and 2.1\,km of real-world robot traversal. GlassGuard achieves 85\% of total glass coverage for its panoramic version. Under identical pinhole inputs, GlassGuard achieves 82\% total coverage, compared with at most 61\% for the evaluated baselines, while producing 5--17$\times$ fewer false voxels per frame. Qualitative examples with a navigation planner illustrate the reconstructed planes blocking paths through glass while leaving traversable routes open. The project page is available at \url{https://glassguardproject.github.io/}.
\end{abstract}

\begin{IEEEkeywords}
Mapping, Sensor Fusion, Range Sensing, Vision-Based Navigation.
\end{IEEEkeywords}
\section{Introduction}

Mobile robots commonly rely on LiDAR-based SLAM for autonomous movement.
However, transparent surfaces remain a fundamental challenge for light-based
sensing: laser energy may be transmitted through glass, specularly reflected,
or only weakly returned, causing ghost points or missed detections. This problem
is particularly relevant to widely used compact LiDARs, which operate at
905\,nm~\cite{livox2024mid360}. Returns from glass
are highly variable and unreliable~\cite{tibebu2021lidar,zhou2024lidar},
potentially producing incomplete or corrupted maps that compromise scene
reconstruction, exploration, and safe navigation.

One way to address this is to exploit physical cues such as return intensity,
reflection patterns, or additional sonar.
Although effective, this route depends on the returns, viewing
geometry, and sensor measurements that happen to be observable, or on
additional hardware. A more broadly accessible alternative is to combine visual glass
detection with surrounding 3D geometry. 

Current efforts can already localize
glass in an image, but converting that detection into a correctly
placed 3D surface remains challenging. This step can be accomplished with a learning-based or a
pipeline-based strategy. A learning-based strategy directly infers glass depth or
plane parameters from observed RGB, and sometimes depth. However,
architectural glass offers almost no features of its own: what a camera sees at
a pane is predominantly the scene behind or reflected in it, which makes the
mapping from image appearance to metric depth ambiguous. Errors in the predicted depth or orientation can
therefore place obstacles in traversable space. 

A pipeline-based strategy instead anchors image evidence, such as detections or
learned depth priors, to nearby metric RGB-D or LiDAR measurements. But,
placement requires associating glass with the correct structure: an apparent
frame or edge may belong to an occluder, or structure behind the glass
(Fig.~\ref{fig:problem}). Both strategies can therefore produce plausible but
incorrectly placed planes in complex scenes. We therefore introduce GlassGuard, an onboard, VRAM-efficient framework that jointly
addresses the metric placement of architectural glass while checking each
proposed plane against the evidence obtained from the image. Rather than relying on an
end-to-end model to generalize directly to 3D glass geometry, GlassGuard shifts
the learned component to the more reliable task of 2D glass detection using the
foundation model SAM3. It then constructs metric 3D candidates from structural
line features extracted from surrounding LiDAR points, reducing the influence of
noisy or spurious returns, and verifies their orientations using depth-free
projective geometry before admitting them to the map. Accepted observations are
consolidated into persistent world-frame planes that can be reinforced, trimmed,
or removed.
The resulting map is designed to cover physical glass while limiting false
occupancy in traversable space.

We evaluate GlassGuard on a Mecanum-wheeled robot across nine building-scale
scenes, six indoor and outdoor environments, and 2.1 km of robot
traversal. Under matched pinhole inputs, GlassGuard achieves higher glass
coverage and substantially less false occupancy than the evaluated
learning-based and pipeline-based baselines, while its panoramic variant further
improves early coverage.

\section{Related Work}

\textbf{Physical-Cue-Based Glass Perception.}
Early robotic systems addressed laser-scanner failures around mirrors and
windows or adapted localization to glass-walled environments
\cite{yang2008dealing,kim2016localization}. Subsequent methods exploit
glass-specific physical cues such as return intensity, reflection patterns,
anomalous or multipath measurements, and view-dependent reflectance
\cite{tibebu2021lidar,foster2023reflectance,zhou2024lidar,weerakoon2024topgn}.
Several systems further incorporate detected glass or reflective planes into
local and global SLAM maps
\cite{wang2017detecting,weerakoon2022cartographer,li2024reflection}. Other
approaches confirm suspected glass through physical contact
\cite{chen2025active} or fusion of time-of-flight and ultrasonic sensing
\cite{hopkins2025realtime}. These methods depend on observable sensor cues,
surface interaction, or additional sensing hardware. Our work considers the
complementary setting in which reliable measurement and additional sensing
hardware cannot be assumed available.

\textbf{2D Glass Detection.}
Visual glass perception has advanced substantially over the past decade.
Representative works include monocular mirror and glass segmentation
\cite{yang2019mirror,mei2020glass}, reflection- and semantic-context modeling
\cite{lin2021rich,lin2022semantic}, polarization-based perception
\cite{kalra2020polarization,mei2022polarization}, and boundary-aware or
progressive refinement \cite{he2021eblnet,yu2022progressive}. More recent
methods exploit reciprocal feature evolution, ghosting cues, and
foundation-model or diffusion priors
\cite{fan2023rfenet,yan2025ghosting,hao2025gem,li2025glasswizard}. These
advances make image-space glass localization increasingly reliable, but 2D
segmentation alone does not provide the metric 3D location at which a detected
surface should be inserted into a navigation map.

\textbf{Visual-to-3D Glass Reconstruction.}
Methods that reconstruct glass from visual input are the closest to our
setting and serve as our baselines. GW-Depth is a supervised monocular estimator that combines glass-structure and
reflection contexts \cite{liang2023gwdepth}; MonoGlass3D, which we evaluate as
the representative learned baseline, jointly predicts glass masks and plane
parameters \cite{zhang2025monoglass3d}. As the representative pipeline-based baseline,
GlassRecon \cite{zheng2026enhancing} (Zheng \emph{et al.}) aligns a
foundation-model depth prior with metric RGB-D measurements using local
RANSAC; for brevity we refer to this pipeline as GlassRecon, the name of its
released dataset. These methods place each surface from a single frame without
checking its orientation or depth, and never revise it afterwards.
\textbf{GlassGuard} instead uses 2D image geometry and persistent plane
management to verify and update the reconstructed glass.

\section{Problem Formulation}
\label{sec:problem}

For robot navigation, glass reconstruction must satisfy two basic occupancy requirements: physical glass should be represented as occupied, while surrounding traversable space should remain free; these requirements motivate our problem formulation. Let $G_t\subset\mathbb{R}^3$ denote the true glass surfaces within the region
observed by time $t$, and let $S_t$ contain all physical surfaces in that
region. The navigation map discretizes this region into voxels $\mathcal{V}_t$
of side length $\rho$ and assigns each voxel $M_t(v)$ as occupied or free.
A glass-aware map is correct at resolution $\rho$ if, for
every voxel $v$,
\begin{equation}
\underbrace{v\cap G_t\neq\emptyset
\Rightarrow
M_t(v)=\mathrm{occ}}_{\text{no miss (safety)}}
\;\wedge\;
\underbrace{v\cap S_t=\emptyset
\Rightarrow
M_t(v)=\mathrm{free}}_{\text{no spill (usability)}}.
\label{eq:correct}
\end{equation}
The first condition requires physical glass to appear as an obstacle, whereas
the second prevents glass reconstruction from contaminating navigable free
space. We call free space that the map wrongly marks as occupied
\emph{spill}, and measure it in Sec.~\ref{sec:experiments}.

GlassGuard targets this voxel-level objective by producing a set of bounded planar segments $\mathcal{P}_t=\{P_i\}$. Their union forms
the reconstructed glass surface, whose intersected voxels are inserted into the
navigation map:
\begin{equation}
\widehat{G}_t=\bigcup_{P\in\mathcal{P}_t}P,
\qquad
\mathcal{V}(\mathcal{P}_t)
=\left\{v\in\mathcal{V}_t\mid v\cap\widehat{G}_t\neq\emptyset\right\}.
\label{eq:planeset}
\end{equation}

\begin{figure}[t]
\centering
\includegraphics[width=0.8\columnwidth]{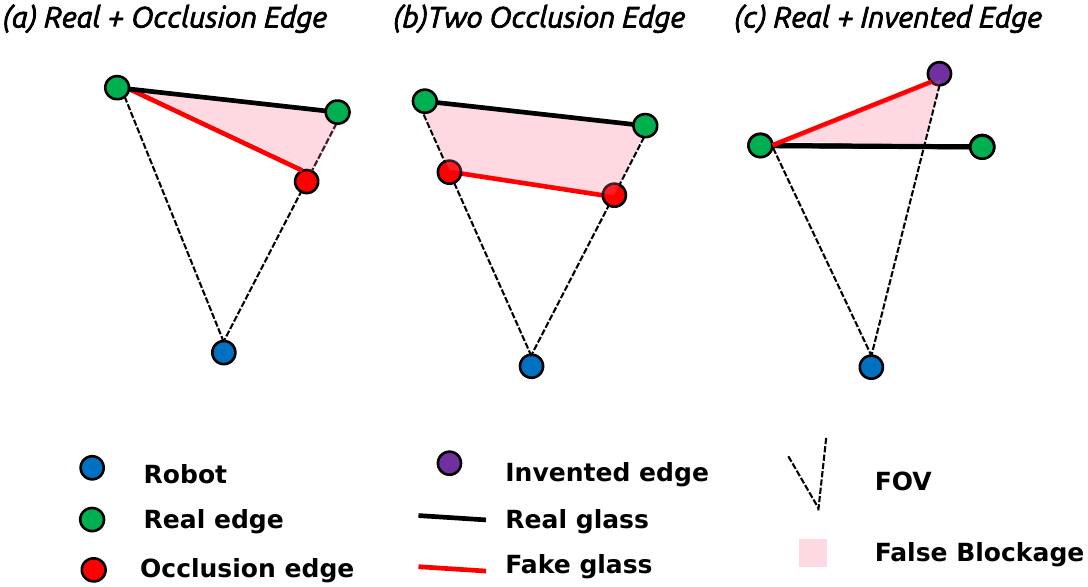}
\caption{Top-down candidate-plane failures: (a) slanted placement from a false
edge, (b) incorrect depth, and (c) reflection-induced slant. Shading denotes
the \emph{false blockage}: navigable free space blocked by the wrong plane.}
\label{fig:problem}
\end{figure}

\section{Method}
\subsection{Overview}
\begin{figure*}[t]
\centering
\includegraphics[width=\textwidth]{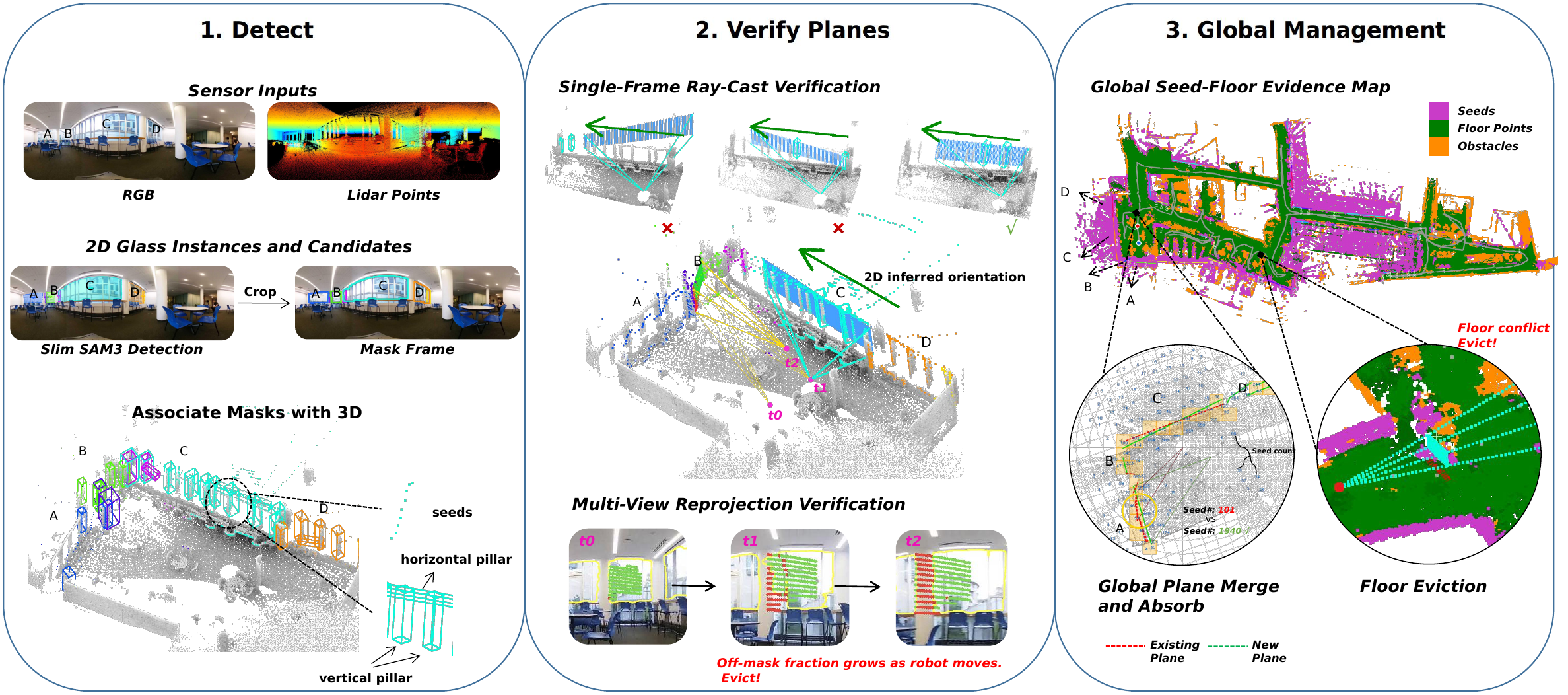}
\caption{System overview; A--D mark four glass instances traced end to end.
\textbf{1.~Detect} (left): \emph{Slim SAM3 Detection} yields instance masks on
the RGB panorama, each cropped to its \emph{Mask Frame}; \emph{Associate Masks
with 3D} projects the LiDAR cloud through that frame to obtain seeds, from which
vertical and horizontal pillars and bounded plane candidates are built
(Sec.~\ref{sec:pillars}). \textbf{2.~Verify Planes} (center): \emph{Single-Frame
Ray-Cast Verification} checks each candidate against the depth-free 2D inferred
orientation (Sec.~\ref{sec:raycast}; crosses rejected, check accepted), and
\emph{Multi-View Reprojection Verification} removes a plane whose off-mask
fraction grows as the robot moves. \textbf{3.~Global Management} (right): the
\emph{Global Seed-Floor Evidence Map} drives \emph{Floor Eviction}, and
\emph{Global Plane Merge and Absorb} resolves overlapping hypotheses by seed
count (Sec.~\ref{sec:global}); surviving planes enter the navigation map.}
\label{fig:pipeline}
\end{figure*}

GlassGuard organizes detection as three stages in
which each module contributes only what it is reliable for
(Fig.~\ref{fig:pipeline}). \emph{First}, a quantized Slim SAM3 student detects
glass-instance masks (Sec.~\ref{sec:slim}); each mask is associated with the
surrounding LiDAR geometry to extract seed points (projected 3D points used to localize glass surfaces). Selected seeds are then grouped into vertical and
horizontal pillars, from which bounded 3D plane candidates are constructed
(Sec.~\ref{sec:pillars}). \emph{Second}, the same mask supplies a depth-free
orientation reference against which the metric 3D candidates are checked:
single-frame ray-cast verification rejects candidates with inconsistent
orientations (Sec.~\ref{sec:raycast}).
\emph{Third}, a global hypothesis manager merges or absorbs overlapping plane
hypotheses using accumulated seed support and inserts the consolidated bounded planes into the
navigation map as occupied space (Sec.~\ref{sec:global}). 
This combination preserves glass coverage while limiting the contamination
of navigable free space by erroneous planes---the two conditions
of \eqref{eq:correct}.

\subsection{Reliable 2D Instance Detection: Slim SAM3}
\label{sec:slim}

Reliable instance proposals are required because every downstream stage is
conditioned on the detected glass mask. SAM3 \cite{carion2026sam3} provides strong
glass/window proposals, but its GPU memory footprint constrains onboard
deployment. We therefore specialize it to glass-only inference using
confidence-guided structural pruning.

Following first-order Taylor pruning \cite{molchanov2017pruning}, channel $j$ is
ranked by
\begin{equation}
I_j=\mathbb{E}\!\left[\left|h_j\frac{\partial L_{\mathrm{conf}}}{\partial h_j}\right|\right],
\label{eq:importance}
\end{equation}
with
\begin{equation}
L_{\mathrm{conf}}=\sum_i \sigma(z_i^{\mathrm{mask}})\,\sigma(z_i^{\mathrm{pres}}),
\label{eq:confidence}
\end{equation}
so importance is measured from the model's own glass confidence, without manual
pruning labels. Low-importance MLP channels and attention heads are structurally
removed, after which the original SAM3 model briefly distills the slim student.

The student retains most of the teacher's glass accuracy
(Table~\ref{tab:glass_seg}): IoU $0.748/0.671$ on GDD/GSD-S versus $0.815/0.713$
for SAM3. Recall, the metric our pipeline depends on, is preserved on GSD-S
($0.918$, the best in the table) and decreases on GDD ($0.895$ to $0.826$).
Grounded-SAM2 and YOLO-World, which produce text-prompted instance masks in
the same way, reach only $0.642$ and $0.436$ IoU on GDD, well below both the
teacher and the student.
Further, with cached text embeddings, the slim student BF16 model and INT8 quantization reduce reserved VRAM by
approximately one third relative to the original SAM3 deployment.

\begin{table}[t]
\centering
\caption{Glass instance segmentation on GDD and GSD-S at $512^2$.}
\label{tab:glass_seg}
\setlength{\tabcolsep}{4pt}
\begin{tabular}{l c c c c c c}
\toprule
 & \multicolumn{3}{c}{GDD} & \multicolumn{3}{c}{GSD-S test} \\
\cmidrule(lr){2-4}\cmidrule(lr){5-7}
Method & IoU & Prec. & Rec. & IoU & Prec. & Rec. \\
\midrule
Grounded-SAM2~\cite{ren2024groundedsam} & 0.642 & 0.781 & 0.783 & 0.509 & 0.563 & 0.841 \\
YOLO-World~\cite{cheng2024yoloworld} & 0.436 & 0.677 & 0.550 & 0.230 & 0.589 & 0.274 \\
SAM3 (original) & \textbf{0.815} & \textbf{0.901} & \textbf{0.895} & \textbf{0.713} & \textbf{0.764} & 0.915 \\
Slim-2816 (ours) & 0.748 & 0.888 & 0.826 & 0.671 & 0.714 & \textbf{0.918} \\
\bottomrule
\end{tabular}
\end{table}

\subsection{Associate 2D to 3D: Pillar Construction}
\label{sec:pillars}
Fitting planes using raw LiDAR points is unreliable because returns are sparse and often hit the cases in Figure 1. We instead construct
candidates from structural line evidence associated with glass frames.

Starting from LiDAR seeds associated with the glass mask, we extract two structural
primitives. \emph{Vertical pillars} are cells occupied across several consecutive
height layers, while \emph{horizontal pillars} are dense seed clusters with a strong
line-like PCA response in the horizontal plane. These structures capture common
glass-frame elements and provide more stable evidence than direct fitting to sparse
returns.

The detected structures are used to form metric plane candidates. Two vertical
pillars determine a candidate direction, whereas a horizontal pillar provides the direction
directly from its axis. 

\subsection{Ray-Cast Orientation Verification}
\label{sec:raycast}

\begin{figure}[t]
\centering
\includegraphics[width=\columnwidth]{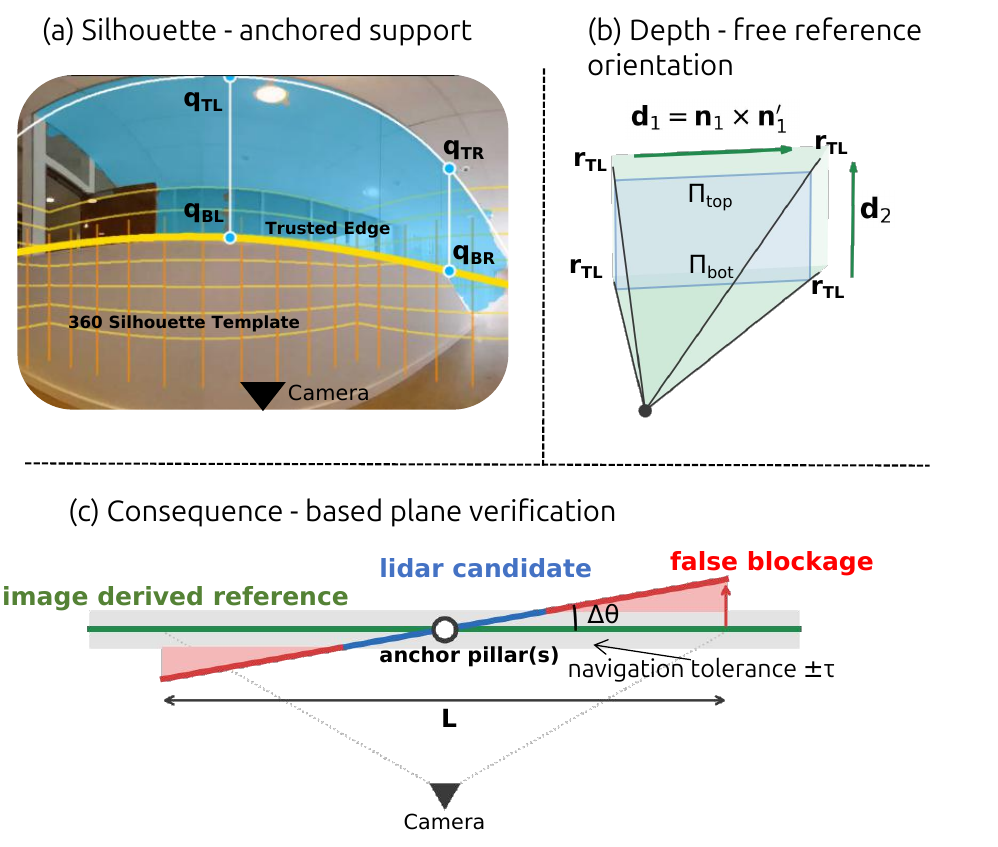}
\caption{Depth-free 2D$\to$3D verification. (a) A reliable silhouette edge
anchors a support quadrilateral enclosing the observed glass mask; its boundary
samples are back-projected as camera rays. (b) Opposite-side interpretation
planes recover the pane's in-plane directions and hence a depth-free reference
orientation. (c) The resulting yaw reference is compared with the LiDAR-derived
candidate; angular disagreement produces a false blockage that widens along the
plane, motivating a length-dependent acceptance threshold.}
\label{fig:raycast}
\end{figure}

A glass mask may contain multiple structural pillars,
and not every combination produces the correct 3D plane
(Fig.~\ref{fig:pipeline}, \emph{Single-Frame Ray-Cast Verification}). The key observation
of this paper is that a 2D image can constrain the possible 3D
directions without requiring depth: two pixels on the line
back-project to rays $\mathbf{r},\mathbf{r}'$, which span an
interpretation plane through the camera center with normal
\begin{equation}
\mathbf{n}=\frac{\mathbf{r}\times\mathbf{r}'}
{\lVert\mathbf{r}\times\mathbf{r}'\rVert}.
\label{eq:interp}
\end{equation}
Any physical 3D line that projects to the observed image line
must lie in this plane. Fig.~\ref{fig:raycast}(a) illustrates how a reliable silhouette edge of the 2D mask determines one interpretation plane. The edge is a robust RANSAC line fit to the mask boundary, and it is discarded or trimmed where its residuals are large.

\textbf{Depth-free orientation recovery.}
A single interpretation plane leaves one degree of ambiguity
because it contains infinitely many possible 3D directions.
That ambiguity is removed when two image lines represent the
same physical in-plane direction---e.g., the top and bottom support
sides, whose interpretation planes are labeled $\Pi_{\mathrm{top}}$
and $\Pi_{\mathrm{bot}}$ in Fig.~\ref{fig:raycast}(b). If their interpretation-plane
normals are $\mathbf{n}_{k}$ and $\mathbf{n}'_{k}$, their
intersection gives the common 3D direction. Repeating this for
two independent direction families determines the plane normal:
\begin{equation}
\begin{aligned}
\mathbf{d}_{k}
&=\frac{\mathbf{n}_{k}\times\mathbf{n}'_{k}}
{\lVert\mathbf{n}_{k}\times\mathbf{n}'_{k}\rVert},
\quad k\in\{1,2\},\\
\mathbf{n}_{\mathrm{ref}}
&=\frac{\mathbf{d}_{1}\times\mathbf{d}_{2}}
{\lVert\mathbf{d}_{1}\times\mathbf{d}_{2}\rVert}.
\end{aligned}
\label{eq:general_orientation}
\end{equation}
No metric depth enters this construction. The recovered normal
is therefore an approximate orientation reference, not an estimate of the
pane's absolute 3D position.
In practice, the dominant navigation error is produced by an
incorrect pane yaw, so full 3D orientation recovery is unnecessary.
For architectural glass, gravity can replace the second image
direction family. Let $\mathbf{n}$ be the interpretation-plane
normal of a reliable horizontal silhouette edge. Its physical
3D direction lies in the interpretation plane and is perpendicular
to gravity $\hat{\mathbf{g}}$, giving
\begin{equation}
\mathbf{d}=(\mathbf{n}\times\hat{\mathbf{g}})\,/\,
\lVert\mathbf{n}\times\hat{\mathbf{g}}\rVert.
\label{eq:gravity_dir}
\end{equation}
Thus, one sufficiently conditioned horizontal edge provides the
yaw reference used by our implementation. 

\textbf{Candidate validation and spatial error.}
The purpose of deriving the image orientation is to detect
possible yaw errors in the LiDAR candidates constructed in the
preceding stage (Fig.~\ref{fig:raycast}(c)). We project both plane normals onto the horizontal
plane as
$\tilde{\mathbf{n}}=\mathbf{n}-(\mathbf{n}^{\top}\hat{\mathbf{g}})
\hat{\mathbf{g}}$, normalize them, and measure their sign-invariant
yaw disagreement:
\begin{equation}
\Delta\theta=\arccos\left|
\hat{\mathbf{n}}_{\mathrm{ref},h}^{\top}
\hat{\mathbf{n}}_{\mathrm{lidar},h}\right|.
\label{eq:residual}
\end{equation}
The absolute value removes the sign ambiguity of a plane normal.
This angular error directly produces the false blockage of
Fig.~\ref{fig:problem}(a) and (c): at distance
$s$ from the rotation pivot, the lateral displacement is
$s\sin\Delta\theta$. Under a centered-pivot approximation, keeping
both ends of a visible pane of length $L$ within spatial tolerance
$\tau$ gives the characteristic scaling
$\Delta\theta\propto\arcsin(2\tau/L)$. Longer panes consequently
require tighter orientation agreement.

Because real candidates may rotate about different structural
anchors, we use Monte-Carlo simulation only to calibrate the
conservative multiplier, which we set to $c=0.75$. The implemented acceptance rule,
fixed across all experiments, is
\begin{equation}
\Delta\theta\le
\operatorname{clamp}\!\left[
0.75\arcsin\!\left(\frac{2\tau}{\max(L,2\tau)}\right),\,
5^{\circ},\,30^{\circ}\right].
\label{eq:tol}
\end{equation}
The spatial tolerance is fixed at $\tau=0.5$\,m, half the navigation voxel.
For $L<2\tau$ the arcsine argument saturates, assigning short panes the
loosest $30^{\circ}$ tolerance. All parameters are fixed across the
experiments. This test verifies orientation consistency; it does not
independently verify candidate depth or metric placement.

\textbf{Metric extent from the accepted candidate.}
The image reference determines orientation and angular support,
but not metric depth. After a LiDAR candidate is accepted, the
left and right support rays are intersected with its metric plane
to obtain the observed pane endpoints. For a candidate ground-plane
line $\mathbf{p}(s)=\mathbf{a}+s\mathbf{d}$, intersection with a
support ray $\mathbf{r}_{i}$ satisfies
$\mathbf{a}+s\mathbf{d}=t_{i}\mathbf{r}_{i}$, giving
\begin{equation}
t_i=\frac{(\mathbf{a}\times\mathbf{d})\cdot\hat{\mathbf{z}}}
{(\mathbf{r}_{i}\times\mathbf{d})\cdot\hat{\mathbf{z}}},
\qquad i\in\{L,R\}.
\label{eq:extent}
\end{equation}
Thus, the image supplies angular support, while the accepted
LiDAR candidate supplies metric placement; their intersection
determines the bounded pane extent.

\subsection{Global Management}
\label{sec:global}

A plane that appears plausible in one frame may be contradicted by later
observations. Instead of permanently inserting every frame-level result, we
maintain each accepted plane as a global hypothesis that can be reinforced,
merged, or removed. The global manager contains three components: plane merging
combines repeated estimates of the same pane, seed--floor evidence removes
planes contradicted by walkable floor, and multi-view verification removes
planes that become inconsistent from new viewpoints.

\textbf{Global plane merging.}
Each frame may produce a new plane estimate for glass that has already been
observed. Because these estimates can differ slightly in orientation, position,
or extent, inserting all of them directly would gradually fill the map with a
cluster of overlapping and misaligned planes. We therefore compare every newly
solved plane with the planes already maintained in the global map.

To determine whether a new plane and an existing track represent the same
glass region, we compare them inside the angular wedge formed by the new
plane's boundary rays. An existing track competes with the new plane when at
least $65\%$ of its ground-plane segment lies inside this wedge, when both
boundary rays intersect it, or when one ray intersects it and their
ground-segment overlap exceeds $70\%$ of the shorter segment. Each track
maintains the total number of supporting LiDAR seeds accumulated over time.
This seed count is the primary support score: the better-supported competitor
survives and absorbs the extent of the weaker estimate. Vertically separated
tracks do not compete because they may represent distinct structures.

\textbf{Seed--floor evidence.}
A false glass plane may cross floor that later observations identify as
walkable, but a single floor observation is insufficient because real glass
frequently reaches the ground. We therefore maintain competing floor and
 glass-seed evidence in a $0.1$\,m ground grid
(Fig.~\ref{fig:pipeline}, \emph{Global Seed-Floor Evidence Map}): an
observation adds one count to its supported class and subtracts one from the
opposing class, with both counts clamped to $[0,3]$. A plane falling on a voxel dominated by floor points is removed, since that voxel should be a walkable region. 

\textbf{Multi-view verification.}
A correctly placed plane should remain aligned with the detected glass region
when observed from different robot poses. For each re-observation we define
the \emph{off-mask fraction} as the fraction of the plane's visible image
projection lying outside all current glass masks
(Fig.~\ref{fig:pipeline}, \emph{Multi-View Reprojection Verification}). Only
well-observed tracks are judged. A correctly placed plane still shows a small
off-mask fraction because masks under-segment the glass, so each track takes the
median of its first two judgments as its own tolerance. An off-mask fraction that
later exceeds this tolerance by more than $0.10$ and is still rising adds one
to an evidence counter, a falling one subtracts one, and the track is evicted
when the counter reaches two; an off-mask fraction of at least $0.55$ evicts it
immediately.
This test addresses
incorrect-depth planes that can pass the single-view yaw check
(Fig.~\ref{fig:problem}(b)): such a plane may align in its original view, but
its projection separates from the glass mask as the viewpoint changes.

\textbf{Planner integration.} Each cycle, the \emph{current} track set is
rendered as dense wall points and inserted into the scans consumed by terrain
analysis and the planners. Nothing persists inside the planner itself: as the
glass map evolves, so does the planner's input. A placed plane therefore blocks
paths even after leaving the field of view, and a removed plane disappears from
the collision map on the next cycle (Sec.~\ref{sec:experiments}).

\section{Experiments}
\label{sec:experiments}

\subsection{Real-World Setup}

We evaluate GlassGuard and the baselines using a Mecanum-wheel robot equipped
with a Livox Mid-360 LiDAR operating at $10$\,Hz
\cite{livox2024mid360} and a $1920\times640$ Ricoh Theta panoramic camera.
At each frame, every LiDAR-using method receives the same $5$\,s stack of
registered point clouds together with synchronized RGB input. The pinhole methods receive the
same pinhole RGB view, while GG-360 receives the full $360^{\circ}$ panoramic
image. GlassGuard runs onboard at a mean of approximately $0.74$\,s per frame on an
NVIDIA RTX 5070 Ti laptop GPU. Its reconstructed planes are inserted into a
LiDAR autonomy stack that provides state estimation, terrain analysis, local
planning, and TARE exploration
\cite{cao2022autonomous,cao2021tare}.

We recorded nine scenes across six indoor and outdoor environments under
daytime and nighttime conditions, totaling more than one hour and $2.1$\,km
of robot traversal. The scenes contain curtain walls, glass doors, office partitions, frosted
panels, glass balustrades, and tinted exterior windows, under sunlight,
reflections, and nighttime lighting.
Planes are placed within $10$\,m of the robot, or $20$\,m in the two large
outdoor scenes.
For evaluation, each method is replayed independently
on the recorded rosbags using the original timestamps, simulating its onboard
execution. The resulting timestamped predictions are then evaluated against
the same manually annotated ground truth.

\subsection{Evaluation Metrics}
We evaluate glass coverage and false occupancy on a common voxel grid fixed in
the shared world frame.

\textbf{Warning-distance coverage.}
Let $D_k(v)$ be the Euclidean distance between the robot and the center of GT
glass voxel $v$ at frame $k$. Voxel $v$ is detected by distance $d$ if it is
occupied in at least one frame where $d\le D_k(v)\le 13$\,m and is line-of-sight visible. Visibility is determined by
ray tracing against observed non-glass surfaces on a $0.2$\,m grid. GT glass beyond the sensing
range is excluded from coverage.
Coverage is
\begin{equation}
C(d)=
\frac{\bigl|\{v\mid v\text{ is detected by }d\}\bigr|}
     {\bigl|\{v\mid v\text{ is visible at some }D_k(v)\ge d\}\bigr|}.
\label{eq:coverage}
\end{equation}
For pinhole methods, eligibility additionally requires the voxel to enter the
camera frustum while the robot is at least $d$ meters away. \emph{Ever}
coverage reports detection in any eligible frame, regardless of distance.

\textbf{Current-map coverage and retention.}
Current-map coverage measures how much nearby glass is represented in the map
at each frame. We use the same eligibility rules as \eqref{eq:coverage}. A voxel
counts as covered if the current map marks it or an adjacent cell as occupied,
allowing a one-voxel tolerance. A previous detection counts only while it
remains in the map. Thus, both delayed detection and later removal reduce
coverage for each affected frame.

We report coverage separately for the collision-critical $0$--$2$\,m and
$2$--$5$\,m bands. For each band, we sum the covered voxel counts across frames
and divide by the summed eligible voxel counts. Retention uses the same
calculation but evaluates each voxel only in frames after it was first
correctly covered within $10$\,m of the robot. This measures whether mapped
glass remains represented, separately from how quickly it is first detected.

\textbf{Spill.}
A predicted voxel is spill if it contains neither active GT glass nor an
observed physical surface. This is the no-spill condition of
\eqref{eq:correct}, with $S_t$ approximated by the surfaces actually observed
during the run. Ground spill lies within $0.25$\,m of the estimated
floor; all remaining spill is air spill. Spill is charged only while the robot
is within $10$\,m of the false voxel (the placement range). The metric spill/frame is the mean number of these false
voxels within the horizon at each frame. Air and Ground report the distinct false voxels accumulated
over the run, together with their median distance to the nearest GT glass
surface. Methods without a persistent map
(MonoGlass3D, GlassRecon) are evaluated on their accumulated detections.

\begin{table}[t]
\centering
\caption{Inputs, peak VRAM, and mean per-frame inference time of the evaluated
methods on an RTX 5070 Ti laptop GPU.}
\label{tab:methods}
\setlength{\tabcolsep}{3pt}
\begin{tabular}{l l c c c c}
\toprule
Method & Camera & LiDAR & Persist.\ map & VRAM & s/frame \\
\midrule
MonoGlass3D & pinhole RGB & --- & --- & 2.4\,GB & \textbf{0.13} \\
GlassRecon  & pinhole RGB & \checkmark & --- & \textbf{0.8\,GB} & 3.6 \\
GG-pin (ours) & pinhole RGB & \checkmark & \checkmark & 1.7\,GB & 0.74 \\
GG-360 (ours) & 360$^\circ$ RGB & \checkmark & \checkmark & 1.7\,GB & 0.73 \\
\bottomrule
\end{tabular}
\end{table}

\begin{table*}[t]
\centering
\caption{Left: nine-scene occupancy comparison at 1.0\,m resolution (coverage in
percent; spill = distinct air/ground false voxels accumulated over the run, mean
concurrent spill per frame, median distance to GT glass). Right: metric sensitivity---Ever coverage /
spill per frame at three further voxel resolutions (tolerance scales with the
voxel); a GlassGuard variant leads every metric at every resolution.}
\label{tab:occupancy}
\setlength{\tabcolsep}{4pt}
\begin{tabular}{@{}c@{\hspace{10pt}}c@{}}
\begin{tabular}[t]{l c c c c c c c c}
\toprule
 & \multicolumn{4}{c}{Coverage $\uparrow$}
 & \multicolumn{4}{c}{Spill $\downarrow$} \\
\cmidrule(lr){2-5}\cmidrule(lr){6-9}
Method & by 10\,m & by 8\,m & by 5\,m & Ever
       & Air & Ground & per frame & med.\ dist \\
\midrule
MonoGlass3D
& 11.8 & 18.6 & 34.6 & 61.0
& 1564 & 490 & 290.2 & 4.2\,m \\
GlassRecon
& 16.6 & 21.0 & 28.0 & 44.4
& 408 & 150 & 85.4 & 3.4\,m \\
GG-pin (ours)
& 32.6 & 54.2 & 68.8 & 82.1
& \textbf{164} & 94 & 16.7 & 3.2\,m \\
GG-360 (ours)
& \textbf{47.6} & \textbf{58.0} & \textbf{71.4} & \textbf{85.2}
& 171 & \textbf{55} & \textbf{15.3} & \textbf{3.0\,m} \\
\bottomrule
\end{tabular}
&
\begin{tabular}[t]{l cc cc cc}
\toprule
 & \multicolumn{2}{c}{0.2\,m} & \multicolumn{2}{c}{0.5\,m} & \multicolumn{2}{c}{2.0\,m} \\
\cmidrule(lr){2-3}\cmidrule(lr){4-5}\cmidrule(lr){6-7}
 & Ever & sp/f & Ever & sp/f & Ever & sp/f \\
\midrule
 & 22.2 & 10452 & 43.4 & 1461 & 80.1 & 52 \\
 & 14.4 & 2370 & 30.1 & 366 & 58.9 & 19 \\
 & 52.0 & \textbf{562} & 72.5 & \textbf{76} & \textbf{90.3} & 6 \\
 & \textbf{60.1} & 700 & \textbf{79.5} & 83 & 89.5 & \textbf{4} \\
\bottomrule
\end{tabular}
\end{tabular}
\end{table*}

\begin{table}[t]
\centering
\caption{Nine-scene current-map protection (pooled over demanded voxel-frames)
at $1.0$\,m and $0.5$\,m voxel resolution, and concurrent spill at $1.0$\,m.
$^{\dagger}$Baselines are evaluated on their accumulated detections, which are
never removed; their retention is 100\% by construction and therefore not
comparable.}
\label{tab:current}
\setlength{\tabcolsep}{2.5pt}
\scriptsize
\begin{tabular}{l ccc ccc c}
\toprule
 & \multicolumn{3}{c}{$1.0$\,m} & \multicolumn{3}{c}{$0.5$\,m} & \\
\cmidrule(lr){2-4}\cmidrule(lr){5-7}
Method & 0--2\,m & 2--5\,m & Ret. & 0--2\,m & 2--5\,m & Ret.
       & Spill/frame $\downarrow$ \\
\midrule
MonoGlass3D & \textbf{98.1} & \textbf{95.5} & ---$^{\dagger}$ & \textbf{84.2} & 81.5 & ---$^{\dagger}$ & 290.2 \\
GlassRecon  & 83.1 & 73.4 & ---$^{\dagger}$ & 63.9 & 58.4 & ---$^{\dagger}$ & 85.4 \\
GG-pin (ours) & 84.4 & 89.0 & 91.7 & 70.1 & 81.0 & 86.8 & 16.7 \\
GG-360 (ours) & 89.1 & 91.5 & \textbf{93.2} & 81.3 & \textbf{83.8} & \textbf{87.8} & \textbf{15.3} \\
\bottomrule
\end{tabular}
\end{table}
\begin{figure*}[t]
\centering
\includegraphics[width=0.75\textwidth, trim=223 42 224 30, clip]{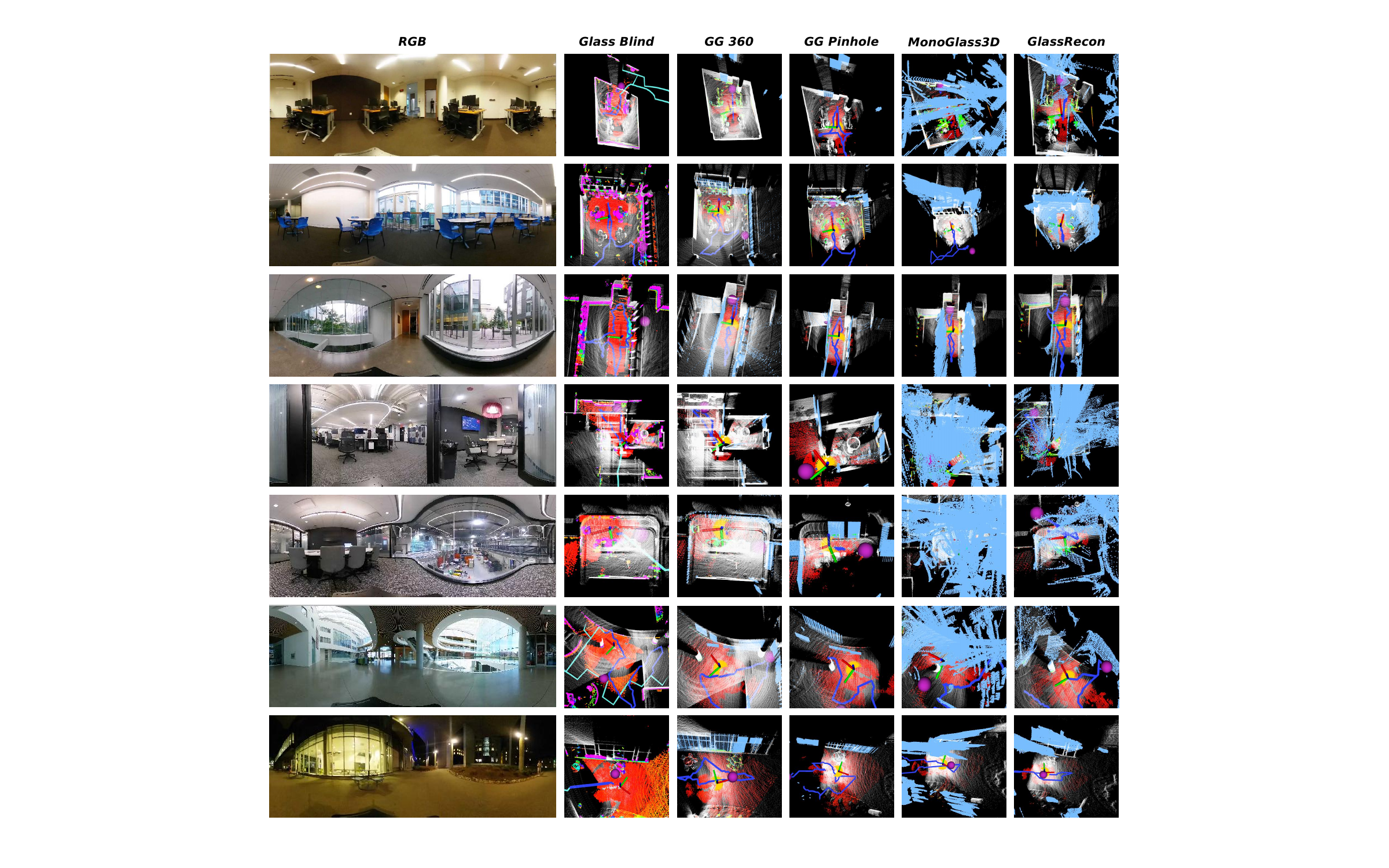}
\caption{Glass mapping and local planning across indoor, outdoor, daytime, and
nighttime scenes. Columns: RGB, Glass Blind, GG-360, GG-pin, MonoGlass3D,
and GlassRecon; azure denotes the local planned path, cyan the global path, and
medium blue the predicted glass.
Without glass detection, planned paths cross physical glass boundaries in rows
3 and 6, potentially directing the robot toward a collision. Both GG variants
recover the glass barriers while largely preserving traversable space.
MonoGlass3D and GlassRecon exhibit substantial false occupancy, particularly in
rows 1, 4, 5, and 6, and leave gaps in glass barriers in some scenes; under
severe map contamination the displayed local paths disappear or become
substantially shorter.}
\label{fig:realvis}
\end{figure*}

\subsection{Quantitative Analysis}
Table~\ref{tab:methods} compares the sensor inputs and computational
requirements of all methods. GG-pin and GlassRecon both use pinhole RGB and
LiDAR, but GG-pin averages $0.74$\,s per frame compared with $3.6$\,s for
GlassRecon, about $5\times$ faster.

As shown in Table~\ref{tab:occupancy}, GG-pin provides the controlled
matched-FOV comparison and outperforms both pinhole baselines at every
warning-distance threshold. It reaches 82.1\% Ever coverage, compared with
61.0\% for MonoGlass3D and 44.4\% for GlassRecon. GG-360 further increases
Ever coverage to 85.2\%. Both GlassGuard variants also detect glass at greater
distances than the baselines: GG-360 already covers 47.6\% of eligible glass
by 10\,m.

The coverage gain is accompanied by substantially less false occupancy.
GG-pin carries 16.7 false voxels per frame, compared with 85.4 for
GlassRecon and 290.2 for MonoGlass3D; GG-360 carries 15.3. Under the same
pinhole input, the baselines maintain $5$--$17\times$ more concurrent false
occupancy than GG-pin.

Table~\ref{tab:current} evaluates the map published to the planner rather than
any-frame discovery. MonoGlass3D attains the highest current coverage, but
only by carrying 290 false voxels per frame. In the
collision-critical bands, GG-360 covers 89.1\% of demanded voxel-frames at
0--2\,m and 91.5\% at 2--5\,m (GG-pin: 84.4\% and 89.0\%). At the finer $0.5$\,m
grid every method loses $7$--$19$ points, and
GG-360 remains ahead of GlassRecon by $17$ points at $0$--$2$\,m while
carrying a fraction of its false occupancy. At $1.0$\,m resolution, both
GlassGuard variants retain coverage in more than 90\% of demanded voxel-frames
after initial detection.

\begin{table}[t]
\centering
\caption{Component ablation on four scenes, GG-pin.
Each row removes one component from the full system; the
two baselines are evaluated on the same four scenes.
Entries are mean $\pm$ standard deviation over the four scenes
($n{=}4$; the deviation indicates scene-to-scene dispersion, not a confidence
interval). $^{\dagger}$Retention approaches 100\% by construction when little or
nothing is ever removed, as for the accumulating baselines of
Table~\ref{tab:current}.}
\label{tab:ablation}
\setlength{\tabcolsep}{1.2pt}
\scriptsize
\begin{tabular}{l c c c c c}
\toprule
Config. & Ever $\uparrow$ & Cur.@0--2\,m $\uparrow$
              & Retention $\uparrow$ & Spill/fr.\ $\downarrow$ & Acc.\ spill $\downarrow$ \\
\midrule
Full GlassGuard        & 85.2$\pm$4.6 & 91.7$\pm$7.6 & 92.2$\pm$4.1 & 19.4$\pm$7.9 & 269$\pm$83 \\
\;-- angle gate        & 86.1$\pm$4.8 & 97.0$\pm$4.9 & 93.2$\pm$3.5 & 34.0$\pm$5.2 & 607$\pm$377 \\
\;-- multi-view & 82.0$\pm$8.4 & 97.1$\pm$5.5 & 93.0$\pm$4.0 & 18.7$\pm$5.3 & 300$\pm$102 \\
\;-- floor evidence    & 82.2$\pm$8.0 & 97.1$\pm$5.9 & 96.2$\pm$0.5 & 25.8$\pm$3.5 & 280$\pm$115 \\
\;-- plane merging     & 84.4$\pm$9.5 & 96.7$\pm$4.6 & 96.8$\pm$1.8$^{\dagger}$ & 33.8$\pm$7.7 & 371$\pm$116 \\
\;-- global mgmt. & 88.2$\pm$6.9 & 99.0$\pm$2.1$^{\dagger}$ & 99.4$\pm$0.4$^{\dagger}$ & 87.7$\pm$27.2 & 478$\pm$164 \\
\midrule
GlassRecon & 48.0$\pm$28.0 & 92.1$\pm$11.8$^{\dagger}$ & 100.0$^{\dagger}$ & 79.2$\pm$50.7 & 585$\pm$541 \\
MonoGlass3D & 51.8$\pm$12.5 & 100.0$\pm$0.1$^{\dagger}$ & 100.0$^{\dagger}$ & 291.4$\pm$56.0 & 2059$\pm$700 \\
\bottomrule
\end{tabular}
\end{table}

\subsection{Component Ablation}
Table~\ref{tab:ablation} removes one component at a time from the full system.
Ever coverage stays at about 82--88\% across all configurations, so the
components separate on false occupancy rather than on coverage.

The angle gate contributes substantially on both measures. Without it, concurrent
spill rises by $75\%$ and $126\%$ more distinct false voxels enter the map, on
every scene ($+40\%$ to $+291\%$)---a larger accumulated rise than removing the
entire manager ($+78\%$), while Ever coverage differs by $0.9$ points, well
within the scene-to-scene spread.

Multi-view verification shows the clearest tradeoff, and its benefit is not
uniform across metrics. Enabling it reduces accumulated spill by $10\%$ and
raises Ever coverage by $3.2$ points, but lowers near-field current coverage by
$5.4$ points and slightly increases concurrent spill, from $18.7$ to $19.4$
voxels per frame. The concurrent coverage it costs outweighs the spill it removes, which
makes this a genuine tradeoff rather than a free gain. Yet planes placed at an
incorrect depth that do not touch the ground need this mechanism to be removed;
no other component reaches them, so a user should weigh it for their
application. Nevertheless, the gaps it opens are short: measured on
GG-360 across the nine scenes of Table~\ref{tab:occupancy} at $\rho=1.0$\,m,
their median length is $2$--$5$ frames and $83$--$94\%$ are recovered.

The mechanisms are complementary. Removing
merging, floor evidence, and multi-view verification one at a time changes
concurrent spill by $+74\%$, $+33\%$, and $-4\%$, about $+104\%$ in total,
whereas removing them together raises it by $352\%$. When one mechanism is
absent the others catch much of what it would have removed.

Even without global management, GG-pin covers far more glass than either
baseline under the same pinhole input and insert-only policy ($88.2\%$ Ever
against $48.0\%$ and $51.8\%$). Its concurrent false occupancy is
comparable to GlassRecon ($87.7$ versus $79.2$ voxels per frame) and less than
one third of MonoGlass3D's, while it admits fewer distinct false voxels over
the run than either ($478$ versus $585$ and $2059$). The advantage over prior
work therefore does not rest on global map management alone.

\subsection{Qualitative Analysis}
Figure~\ref{fig:realvis} compares glass mapping and its effect on local
planning across all methods and environment types. Without glass detection,
planned paths cross physical glass boundaries; both GG variants recover the
barriers while largely preserving traversable space, whereas the baselines'
false occupancy extends into traversable regions and, under severe map
contamination, suppresses or shortens the displayed local paths. These
examples illustrate the planning relevance of jointly reducing missed glass
and spill into free space; they demonstrate planner integration but are not
treated as a quantitative navigation-success evaluation. The accompanying
video shows these runs in full.

\section{Conclusion}
In this work, we present GlassGuard, a navigation-oriented framework for
reliable architectural-glass mapping in real-world environments. We approach the
problem from a new angle, from how we formulate the objective, to depth-free 2D
geometric verification, to persistent system-level verification from multiple
sources of global evidence. Extensive
real-world experiments across diverse environments demonstrate higher glass
coverage with substantially less false occupancy than the evaluated baselines.
Room for improvement remains: faster detection and inference would raise
the update rate, and the current planar, near-vertical assumption could be
extended to curved or strongly tilted glass, which we leave to future work. Based on its onboard performance, we
regard GlassGuard as a significant step toward deployable glass-aware robot
navigation.

\end{document}